\documentclass[12pt,a4paper]{article}
\usepackage[a4paper,margin=0.9in]{geometry}
\usepackage{amsmath}
\usepackage{amssymb}
\usepackage{amsfonts}
\usepackage{bm}
\usepackage{amsthm}
\usepackage{algorithm}
\usepackage{algpseudocode}
\usepackage{multirow}
\usepackage{longtable}
\usepackage{tabularx}

\usepackage{graphicx}
\usepackage{subcaption}
\usepackage{cite}
\usepackage{url}
\newtheorem{definition}{Definition}
\usepackage{setspace}
\usepackage[hidelinks]{hyperref}

\begin{document}

\title{\parbox{\textwidth}{\centering
\Large\bfseries
Learning the Mathematical Property for Designing Low Mutual Coherence Binary Sensing Matrices
}}

\author{
\normalsize
Rekha$^{a}$, Santosh Singh$^{a}$, and S. K. Neogy$^{b}$\\[6pt]
\small
\parbox{0.90\textwidth}{\centering
$^{a}$Department of Mathematics, Shiv Nadar Institution of Eminence, Delhi, India.\\
(e-mail: \texttt{re824@snu.edu.in})\\[5pt]
$^{b}$Centre for Research on Economics and Data Analysis (CREDA),\\
Indian Statistical Institute, New Delhi, India. \\
}
}
\date{}

\maketitle
\begin{abstract}
The essential part for the success of almost all learning-based methods relies on the quality and size of the data sets chosen, based on the application and problem. In this research work, we are constructing the sensing matrix, which is essential for the success of the compressive sensing technique. We have chosen a learning-based technique for the construction of the sensing matrix. The novelty and uniqueness of the proposed technique is that it does not use any data set and also does not use a specific application. It uses the mathematical property/constraint for the construction of the sensing matrix for the perfect recovery of the signal. The perfect recovery of signals is an old and still very challenging problem in real-world applications. There have been many techniques for the perfect recovery of signals. In late 2000, compressive sensing became a popular mathematical tool for the perfect recovery of sparse signals. The core of the compressive technique is the construction of the sensing matrix, which satisfies certain special properties such as restricted isometry property (RIP), null space property (NSP), and spark property (SP). All these properties are NP-hard problems and hence computationally challenging to solve. For all practical purposes, the construction of the sensing matrix needs to achieve low mutual coherence to achieve the perfect recovery of the signals.  We have used a neural network for the construction of the sensing matrix, and this framework constructs a binary sensing matrix with low mutual coherence. The entries in the matrix are generated through a shared underlying rule. The proposed architecture is simple and does not use large-scale training data sets. Such uniqueness and novelty bring a drastic reduction in computational cost, and also, for the first time in literature, the use of a mathematical property for defining the loss function. In this proposed research work, the mutual coherence property has been used in the neural network framework. Such a neural network framework brings generality (application dependency), robustness, and reduces storage requirements. 

The simulation results demonstrate the effectiveness of the proposed approach. The results show that the proposed learning-based method consistently outperforms popular conventional methods, which are based on deterministic as well as random-matrix techniques.

\end{abstract}

\vspace{6pt}


\section{Introduction}\label{sec_1}
Compressive sensing (CS) has emerged as an efficient signal acquisition framework that enables the recovery of sparse signals from a reduced number of measurements. Among the various components of compressive sensing, the design of the sensing matrix plays a fundamental role in efficiently acquiring the signal and in determining the accuracy and reliability of signal reconstruction. For successful sparse signal recovery, the sensing matrix must satisfy certain theoretical conditions that have been extensively studied in the literature. Some of the most important properties includes the spark \cite{donoho2003optimally, donoho2006compressed}, Null Space Property (NSP) \cite{donoho2006compressed, cohen2009compressed}, Restricted Isometry Property (RIP) \cite{candes2006robust, candes2005decoding, candes2008restricted}, and Mutual Incoherence Property (MIP) \cite{elad2007optimized, candes2007sparse, bourgain2011explicit}. These properties provide theoretical guarantees for sparse signal reconstruction and have motivated the development of several techniques for constructing efficient sensing matrices.

Among the theoretical conditions used to evaluate sensing matrices, the Restricted Isometry Property (RIP) and the Mutual Incoherence Property (MIP) are the most widely studied. Although the RIP provides strong theoretical guarantees for sparse recovery, verifying whether a sensing matrix satisfies this property is an NP-hard problem \cite{tillmann2013computational}. In contrast, the Mutual Incoherence Property is computationally tractable and can be evaluated more efficiently. Therefore, this paper primarily focuses on the Mutual Incoherence Property because of its practical applicability and analytical convenience. A sensing matrix with low mutual coherence generally enhances the uniqueness and stability of sparse signal reconstruction. David Donoho and Michael Elad \cite{donoho2003optimally} further emphasized the significance of incoherence by showing that sparse recovery performance improves when the sensing matrix $\bm{\Phi}$ is sufficiently incoherent with the sparsifying basis $\bm{\Psi}$. Consequently, the design of sensing matrices with minimum mutual coherence has become an important research direction in compressive sensing.

Over the years, various approaches have been proposed for sensing matrix design, accounting for factors such as computational complexity, storage efficiency, reconstruction accuracy, and adaptability to the underlying signal structure. The commonly studied sensing matrix constructions include random matrices, deterministic matrices, and learning-based approaches. Each category offers distinct advantages and trade-offs, making it suitable for different compressed sensing applications and implementation constraints.

Among these, random sensing matrices generated from Gaussian or Bernoulli ensembles are widely used because they satisfy important theoretical conditions with high probability and provide strong guarantees for sparse signal recovery \cite{candes2008restricted, candes2006stable, baraniuk2008simple, candes2007sparsity}. Random matrices are largely incoherent with most sparsifying bases. However, due to their dense and unstructured nature, such matrices require significant storage space and lead to high computational complexity during signal acquisition and reconstruction \cite{devore2007deterministic, ravelomanantsoa2015compressed}.
 
To address these limitations, deterministic sensing matrix constructions have been investigated extensively \cite{elad2007optimized, duarte2009learning, xu2010optimized, yi2021new, wang2023novel}. These approaches aim to minimize mutual coherence between the sensing and sparsifying matrices. Many of these methods formulate the sensing matrix design problem as an optimization problem in which the corresponding Gram matrix is forced to approximate an ideal target Gram matrix. Such optimization problems are generally solved using iterative algorithms. While these methods often provide improved performance compared to conventional random matrices, their construction procedures remain computationally expensive and mathematically complex \cite{yi2021new, devore2007deterministic}. 

In recent years, deep learning techniques have attracted significant attention in the field of compressive sensing, demonstrating remarkable success in both signal reconstruction and sensing matrix construction \cite{adler2016compressed, wu2019learning, bora2017compressed, monga2021algorithm}. Most learning-based approaches jointly optimize the sensing matrix and reconstruction network within a unified training framework. In these methods, both the sensing and recovery stages are represented as trainable layers of a neural network and are optimized using large training datasets. By learning the underlying structure of signals directly from data, these approaches often achieve superior reconstruction performance. However, they typically rely heavily on extensive training data and computationally intensive architectures. Furthermore, learning-based methods primarily focus on data-driven optimization rather than explicitly enforcing the theoretical properties required for compressive sensing \cite{bora2017compressed, mousavi2015deep, zhang2018ista}. As a result, theoretical guarantees such as incoherence or RIP are often not well established.

Motivated by the limitations of existing sensing matrix construction approaches, this article proposes a novel sensing matrix design framework that primarily aims to learn binary sensing matrices by optimizing coherence-based mathematical properties. As far as we know, this property-based learning approach is novel and not limited to this area; it can be applied across multiple disciplines. The main contributions of this article are summarized as follows:

\begin{itemize}
    \item A novel framework is proposed for constructing sensing matrices with reduced mutual coherence, which is one of the most important properties in compressive sensing.
    
    \item The proposed method generates binary sensing matrices whose entries are constructed according to a common rule-based mechanism, which reduces computational complexity and storage requirements
    
    \item Unlike conventional learning-based approaches that employ deep, computationally intensive architectures, the proposed framework uses a simple neural network to efficiently generate the sensing matrix.
    
    \item The proposed model uses a combined loss function, which favors incoherence.
    
    \item Experimental results demonstrate that the proposed sensing matrix achieves lower mutual coherence compared to several existing sensing matrix construction methods.
\end{itemize}

\section{Background and Preliminaries} \label{sec_6_2}
In the first decade of the 2000s, Compressive Sensing (CS) theory emerged as a modern signal sampling technique. Unlike the traditional Nyquist sampling method, which often requires a large number of measurements, CS enables the acquisition of high-dimensional signals using far fewer samples. Despite this reduced sampling, the original signal can still be accurately reconstructed using suitable reconstruction algorithms. The fundamental idea behind compressive sensing is the sparsity assumption. A signal is considered $k$-sparse if it contains at most $k$ non-zero coefficients in an appropriate representation domain. Therefore, CS assumes that the signal has a sparse representation in a suitable transform domain, enabling efficient sampling and accurate reconstruction.

Assume $\bm{s} \in \mathbb{R}^N$ is the high-dimensional original signal and $\bm{\Psi}\in \mathbb{R}^{N \times N}$ is the sparsifying basis. In the sparsifying domain, the original signal can be represented with only a few coefficients and mathematically can be expressed as,
\begin{equation}
    \bm{s=\Psi x}
\end{equation}
where $\bm{x}\in \mathbb{R}^N$ is the sparse representation of $\bm{s}$. Then, instead of measuring every sample, compressive sensing measures the compressed measurement with the help of the sensing matrix $\bm{\Phi}\in \mathbb{R}^{M \times N}$ where $M\ll N$. Using the sensing matrix, the $N$-dimensional signal is compressed into $M$ measurements, and the measurement process can be represented as, 
\begin{equation}
    \bm{y=\Phi s = \Phi \Psi x = Ax}
\end{equation}
where $\bm{y} \in \mathbb{R}^M$ is the low-dimensional measurements. In the above formulation, the matrix $\bm{\Phi}$ is the sensing matrix, and $\bm{A}$ is popularly known as the effective sensing matrix. 

The success of compressive sensing largely depends on three key components: the sparsifying basis, the sensing matrix, and the recovery algorithm. Considerable research has been devoted to the development and improvement of each of these components. Among them, the sensing matrix plays a particularly important role because it directly influences both signal acquisition and reconstruction. Therefore, the design of an efficient sensing matrix is a fundamental aspect of compressive sensing.

\subsection{Conditions for Perfect Signal Recovery}
An effective sensing matrix should be able to capture the most significant information of a signal using only a small number of measurements while preserving the information required for accurate reconstruction. However, not every matrix possesses these desirable properties. To address this challenge, researchers have proposed several mathematical conditions to enable sensing matrices to efficiently acquire signal information without significant loss, while ensuring reliable signal recovery from the compressed measurements. Among them, the Restricted Isometric Property (RIP) was the most popular sufficient condition, which was introduced by Emmanuel Candès and Terence Tao in 2005 \cite{candes2005decoding, ncandes2006near}.
\begin{definition}[\textbf{Restricted Isometry Property}] \cite{candes2005decoding}
A matrix $\bm{A} \in \mathbb{R}^{M \times N}$, satisfies the Restricted Isometry Property (RIP) of order $k$ if there exists a isometry constant $\delta_k \in (0,1)$ such that
\begin{equation} \label{eq_RIP}
(1 - \delta_k)\|\bm{x}\|_2^2 \leq \|\bm{A} \bm{x}\|_2^2 \leq (1 + \delta_k)\|\bm{x}\|_2^2 .
\end{equation}
holds for all $k$-sparse signals.
\end{definition}
RIP preserves the mutual distances between all sparse signals after mapping to the measurement space, ensuring that no two sparse signals map to the same measurement. If the effective sensing matrix $\bm{A}$ satisfies the restricted isometric property, then this is sufficient for a variety of algorithms to be able to successfully recover the original signal. However, verifying whether a given matrix satisfies RIP is itself NP-hard \cite{tillmann2013computational}. Whereas the Mutual Incoherence Property (MIP) \cite{donoho2003optimally} provides a more practical and trackable approach for constructing the sensing matrix. 

A matrix satisfies MIP if its mutual coherence is sufficiently small, and mutual coherence measures the similarity between the columns of the sensing matrix; the mathematical definition is given below.
\begin{definition}[\textbf{Mutual Coherence}] \cite{davenport2012introduction}\label{def3}
Let $\bm{A}=[a_1,a_2,...a_N],a_i \in \mathbb{R}^M$, then the mutual coherence of $\bm{A}$ is the maximum correlation between the columns of $\bm{A}$, i.e.,
\begin{equation} \label{eq_mcoh}
\mu(\bm{A}) = \mathop {\max\limits_{1 \leq i, j \leq N}}_{i \neq j} \frac{|a_i^T a_j|}{\|a_i\|_2 \|a_j\|_2}.
\end{equation}
\end{definition}

In 2003, David Donoho and Michael Elad \cite{donoho2003optimally}, shows that if sparsity of the signal satisfies
\begin{equation} \label{eqn6}
k < \frac{1}{2} \left(\frac{1}{\mu(\bm{A})} + 1\right), 
\end{equation}
then a $k$-sparse signal can be recovered exactly using either Basis Pursuit or Orthogonal Matching Pursuit (OMP). This result demonstrates that lower mutual coherence between the sensing matrix and the sparsifying basis allows the accurate reconstruction of signals with relatively denser sparse representations within the compressive sensing framework. Consequently, mutual coherence has emerged as a practical and computationally efficient alternative for evaluating and designing sensing matrices. Therefore, this paper proposes a novel approach to constructing a learning-based sensing matrix that aims to achieve lower mutual coherence.

\subsection{Some Important Terminologies}
In CS, accurate sparse recovery is generally achieved when the columns of the effective sensing matrix exhibit low mutual coherence. Therefore, for a given sparsifying basis $\bm{\Psi}$, the objective is to design a sensing matrix $\bm{\Phi}$ such that the coherence among the columns of $\bm{A}$ is minimized. However, directly minimizing the maximum mutual coherence is difficult because of its non-differentiable objective. Therefore, to guide the learning process and to measure the performance of the proposed matrix, coherence and related majors were utilized and described below.

\begin{itemize}
    \item \textbf{Gram Matrix:} For the effective sensing matrix $\bm{A}$ the corresponding Gram matrix is defined as
    \[
    \bm{G} = \bm{A}^T\bm{A}. 
    \]
    The diagonal entries of $\bm{G}$ represent the squared norms of the columns of $\bm{A}$, which become unity when the columns are normalized. The off-diagonal entries represent the pairwise coherence between different columns of $\bm{A}$. Since lower coherence improves sparse reconstruction performance, the off-diagonal elements of $\bm{G}$ are ideally expected to be as close to zero as possible.

    \item \textbf{Maximum Mutual Coherence: } Maximum mutual coherence of $\bm{A}$ measures the largest coherence value between any two distinct columns of the effective sensing matrix. It is defined as
    \[
    \mu(\bm{A})=\max_{i\neq j}\{|g_{ij}|\}
    \]
    where $g_{ij}$ denotes the $(i,j)$-th entry of the Gram matrix $\bm{G}$. A smaller value of $\mu(\bm{A})$ indicates lower worst-case coherence and generally leads to improved sparse recovery guarantees.

    \item \textbf{Average Mutual Coherence: } While maximum mutual coherence considers only the largest pairwise coherence, average mutual coherence evaluates the overall coherence distribution among the columns of the sensing matrix. It is defined as
    \[
    \mu_{\mathrm{avg}}(\bm{A}) = \mathrm{mean}\left(\sum_{i\neq j}|g_{ij}|\right) 
    \]
    A lower value of average mutual coherence implies that, on average, the columns of the sensing matrix are weakly correlated, thereby contributing to more stable and reliable sparse signal reconstruction.

    \item \textbf{Total Mutual Coherence: } Total mutual coherence measures the cumulative coherence among all distinct column pairs of the effective sensing matrix and is expressed as
    \[
    \mu_{\mathrm{total}}(\bm{A})=\sum_{i\neq j}|g_{ij}|
    \] 
    Minimizing this quantity encourages a globally incoherent sensing matrix structure, which is beneficial for robust compressive sensing performance.

    \item \textbf{Equiangular Tight Frames: }  Equiangular Tight Frames (ETF) are particularly important because they attain the lowest possible coherence bound under certain conditions. The definition of ETF is given below.
    \begin{definition}[\textbf{Equiangular Tight Frame}] \cite{sustik2007existence}\label{def_3_3}
        Let $\bm{A} = [a_1,a_2,...a_N]$ be an $M \times N$ matrix. Then $\bm{A}$ is said to be an equiangular tight frame (ETF) if it satisfies the three conditions below
        \begin{enumerate}
        \item $\|a_i\|=1$ for $i = 1,2,...,N$.
        \item There exists a constant $d$ such that, $|\langle a_i, a_j \rangle |=d$ for $i \neq j$, here $\langle a_i, a_j \rangle$ represents the inner product.
        \item The columns of $\bm{A}$ form a tight frame, i.e., $\bm{A}\bm{A}^T = \frac{N}{M} \mathbf{I}$.
        \end{enumerate}
    \end{definition}
   To encourage highly incoherent measurements, ETF-inspired constraints are incorporated into the proposed sensing matrix learning framework. Such a structure promotes uniform coherence distribution and improves the robustness of sparse signal recovery.

    \item \textbf{Binary Sensing Matrices: } A sensing matrix is referred to as a binary sensing matrix when its entries are restricted to two discrete values, typically $\{0,1\}$ or $\{-1, 1\}$. 
    
    Binary sensing matrices are particularly attractive for practical implementations due to their low storage requirements, reduced computational complexity, and hardware-efficient realization. These characteristics make them well-suited for efficient learning and deployment of sensing matrices in real-world compressive sensing systems. However, maintaining incoherence among the columns with this binary constraint is the main difficulty. 
\end{itemize}

\section{Existing Sensing Matrix Design} \label{sec_6_3}
The performance of compressive sensing largely depends on the design of the sensing matrix. An efficient sensing matrix should be able to capture the essential information of a high-dimensional sparse signal with a significantly smaller number of measurements while still enabling accurate signal reconstruction. Therefore, the construction of the sensing matrix remains a central research problem in compressive sensing.

As discussed in the above section, most of the theoretical properties, including the Null Space Property (NSP), the Restricted Isometry Property (RIP), and the Spark property, provide strong recovery guarantees. However, verifying or constructing matrices that strictly satisfy them is computationally difficult and often impractical for large-scale applications. Consequently, researchers have focused on more tractable alternatives, among which mutual coherence has emerged as one of the most widely adopted measures for sensing matrix design.

Based on coherence minimization and practical implementation considerations, sensing matrices can be broadly categorized into random, deterministic, and data-driven learned matrices. Each class offers different trade-offs in terms of recovery performance, computational complexity, storage requirements, and implementation feasibility.

\subsection{Random Sensing Matrix}
Random sensing matrices gained significant attention during the early development of compressive sensing due to their strong theoretical recovery guarantees. In these constructions, matrix entries are typically sampled from random distributions such as Gaussian or Bernoulli distributions. Random matrices satisfy the RIP with high probability and exhibit low mutual coherence with most sparsifying bases \cite{baraniuk2008simple}.

The major advantage of random sensing matrices lies in their universality and strong reconstruction capability. However, despite their theoretical effectiveness, they suffer from several practical limitations, including high storage complexity, large memory requirements, and difficulties in hardware implementation for large-scale systems.

\subsection{Deterministic Sensing Matrix}
To address the practical limitations of random constructions, researchers explored deterministic sensing matrix design strategies. Following the influential work of Michel Elad \cite{elad2007optimized} on coherence optimization, several iterative and algebraic approaches were proposed to construct matrices with progressively reduced mutual coherence \cite{xu2010optimized, yi2021new, wang2023novel}. 

Deterministic matrices offer advantages such as reproducibility, reduced randomness, and improved implementation feasibility. In many practical scenarios, they also demonstrate competitive or superior performance compared to purely random constructions. However, these methods often rely on complex algebraic formulations or computationally intensive iterative optimization procedures, which may reduce flexibility and increase design complexity.

\subsection{Data-driven Learned Sensing Matrix}
Recent advances in machine learning and deep neural networks have significantly influenced compressed sensing research, leading to the development of data-driven sensing and reconstruction frameworks. Unlike conventional approaches that construct sensing matrices based on analytical principles and theoretical guarantees, learning-based methods optimize sensing and reconstruction processes directly from training data to improve performance for specific signal classes.

Initially, deep learning was primarily employed for signal reconstruction, while the sensing stage continued to rely on conventional random sensing matrices. Although these approaches achieved notable improvements in reconstruction quality, they did not address the fundamental limitations associated with random sensing matrices, particularly their high storage requirements and computational overhead. Consequently, the storage burden remained largely unchanged despite advances in reconstruction algorithms.

Mousavi et al. \cite{mousavi2015deep} employed stacked denoising autoencoders (SDAs) to jointly learn the measurement and recovery processes. This is among the first papers to use a neural network to learn a compressive sensing framework. Their framework was trained using the ImageNet dataset \cite{russakovsky2015imagenet} and demonstrated the effectiveness of end-to-end learning for compressed sensing. Nevertheless, the approach relied heavily on extensive training data and computational resources.

Edler et al. \cite{adler2016compressed} later proposed a comprehensive deep learning framework that jointly optimized the acquisition and reconstruction stages. By adapting the sensing process to the data distribution, the method achieved improved reconstruction accuracy compared with conventional approaches. However, as with many learned sensing strategies, theoretical guarantees such as the Restricted Isometry Property (RIP) and low mutual coherence were generally absent.

In 2017, Bora et al. \cite{bora2017compressed} drew attention for introducing a generative-model-based reconstruction framework that exploits the underlying structure of signals. Their approach was trained on extensive datasets, including MNIST, a dataset of 60,000 handwritten digit images \cite{lecun1998gradient}, and CelebA, a dataset of more than 200,000 celebrity face images \cite{liu2015deep}. While the method improved reconstruction performance by leveraging learned data priors, the sensing process remained dependent on data-driven training and large-scale datasets.

A successful deep learning-based compressed sensing framework, ReconNet, proposed by Lohit et al. \cite{lohit2018convolutional}, utilized a random Gaussian sensing matrix and focused exclusively on learning an effective reconstruction network. The model was trained using the large-scale ImageNet dataset \cite{krizhevsky2012imagenet}, demonstrating the potential of deep learning for compressed sensing recovery while retaining conventional random measurements.

Zhang et al. \cite{zhang2018ista} proposed ISTA-Net, a structured deep network inspired by the Iterative Shrinkage-Thresholding Algorithm (ISTA) \cite{beck2009fast}. In this framework, each iteration of the optimization algorithm was unfolded into a neural network layer, and the algorithm parameters were learned directly from data. Trained on the ImageNet dataset \cite{krizhevsky2012imagenet}, the method achieved faster convergence and improved reconstruction performance. However, its effectiveness remained dependent on the availability of large training datasets and substantial computational resources.

A more direct attempt to learn the sensing matrix was made by Shi et al. \cite{shi2019image} in 2019 through the CSNet framework, which jointly optimized sensing and reconstruction. The authors learned three different sensing matrix types: floating-point, binary, and bipolar matrices. Although binary and bipolar sensing matrices reduce storage requirements compared with dense random matrices, the learned matrices do not provide theoretical guarantees such as RIP or mutual incoherence. Furthermore, the framework was trained on the BSD500 dataset \cite{arbelaez2010contour}, requiring a considerable amount of training data. Since the sensing matrix is optimized for a particular data distribution, its performance may not generalize well to signals that differ significantly from the training set.

Overall, learning-based compressed sensing methods have demonstrated impressive reconstruction performance by adapting to the statistical characteristics of specific datasets. However, most existing approaches either continue to employ random sensing matrices, thereby retaining the associated storage and computational burdens, or rely on learned sensing matrices that require large-scale training data and expensive optimization procedures. Moreover, learned sensing matrices generally lack rigorous theoretical guarantees, such as RIP and mutual incoherence, and their performance is often closely tied to the training distribution, limiting their generality and applicability in lightweight, resource-constrained, or deployment-critical environments.

\section{Motivation} \label{sec_6_4}
Although substantial progress has been made in sensing matrix design, existing approaches still face important limitations. Random matrices provide strong theoretical guarantees but require significant storage and are often inefficient for practical implementation. Deterministic constructions improve structural control and reproducibility; however, many of these methods involve complicated algebraic designs or computationally intensive optimization processes. On the other hand, data-driven learned sensing matrices achieve adaptability and improved reconstruction performance but generally depend on deep learning frameworks, large training datasets, and high computational resources.

Consequently, existing methods are unable to simultaneously achieve the desirable properties of low storage complexity, low mutual coherence, simple construction, and lightweight learning. This limitation highlights the need for a more efficient sensing matrix design framework that balances theoretical effectiveness with practical implementation requirements.

Motivated by these challenges, this study introduces a lightweight framework for generating binary sensing matrices via shared neural parameterization. The proposed method is designed to minimize mutual coherence while substantially reducing storage complexity, thereby eliminating the need for deep network architectures and large-scale training datasets. In contrast to conventional learning-based approaches that rely heavily on data-driven structural learning, the proposed framework focuses on learning from the desired sensing properties themselves rather than from training data. This property-oriented learning strategy enhances the generalization performance of the sensing matrix while alleviating the computational and data burdens associated with extensive training. Consequently, the framework offers an efficient, scalable, and practical solution for compressive sensing applications, particularly in resource-constrained environments.

\section{Proposed Property-driven Learned Sensing Matrix Design} \label{sec_6_5}
The performance of a compressive sensing framework is strongly influenced by the properties of the sensing matrix. Conventional random sensing matrices, although widely used, often impose significant storage overhead and present practical challenges for hardware implementation. Existing learning-based approaches aim to overcome these limitations; however, they often rely on large-scale training datasets and computationally intensive deep architectures. Moreover, such methods primarily learn the underlying structure of the training data, which limits their generalization and provides only limited theoretical guarantees regarding the properties of the resulting sensing matrix.

To address these challenges, this work proposes a binary sensing matrix constructed through a property-driven learning framework, in which the sensing matrix is optimized directly based on the mutual incoherence property rather than being learned from signal data. By focusing on the intrinsic theoretical characteristics required for effective compressive sensing, the proposed approach enhances generalization while maintaining practical suitability for efficient storage and hardware implementation.

\subsection{Problem Formulation} 
The primary objective of this chapter is to construct a binary sensing matrix with low coherence, whose entries are generated through a shared underlying rule. From a classical optimization perspective, this problem involves a combinatorial search, making it computationally intractable for large-scale settings. To overcome this challenge, we employ a neural network-based framework. However, unlike conventional deep learning approaches, we avoid highly complex architectures and large-scale training datasets, as they increase computational overhead and complicate the overall design. Instead of learning from signal datasets, our approach aims to construct the sensing matrix by directly learning and optimizing its inherent mathematical property, namely coherence, which is fundamental to compressive sensing. By focusing on the sensing matrix property rather than the data itself, the proposed framework reduces storage requirements while improving its universality, robustness, and practical applicability across diverse signal domains.

Accordingly, we formulate this problem as an optimization problem dependent on the neural network parameters. Let $\bm{A}_{\theta}$ denote the desired learned sensing matrix with parameters $\theta$. The problem can then be formulated as
\begin{equation}
    \begin{aligned}
    \min_{\theta} \quad & \mathcal{L}(\bm{A}_{\theta}) \\
    \text{subject to} \quad & \bm{A}_{\theta}(i, j) \in \left\{\frac{-1}{\sqrt{M}}, \frac{1}{\sqrt{M}}\right\}, \\
    & \text{Columns of}\; \bm{A}_{\theta}\; \text{generated by shared-rule}.
    \end{aligned}
\end{equation}

Here, $\mathcal{L}(\bm{A}_{\theta})$ denotes the loss function that promotes incoherence, which is discussed in the following section.

Our objective is to learn a binary sensing matrix $\bm{A}_{\theta} \in \mathbb{R}^{M \times N}$ such that
\[
a_{ij} \in \left\{\frac{-1}{\sqrt{M}}, \frac{1}{\sqrt{M}}\right\}.
\]

Instead of randomly selecting or directly optimizing the entries $a_{ij}$ of $\bm{A}_{\theta}$, we generate each column of $\bm{A}_{\theta}$ using a neural network governed by a shared rule. Let
\[
z_j \in \mathbb{R}^M,\; \forall \;j=1,2,\dots N
\]
denote the latent vectors distributed according to a predefined distribution, which are provided as inputs to the neural network. Then, a lightweight neural network
\[
\mathcal{N}_\theta:\mathbb{R}^M \to \mathbb{R}^M
\]
maps each latent vector to an intermediate representation, $s_j$, i.e.,
\[
s_j = \mathcal{N}_\theta(z_j)
\]
from which the corresponding binary sensing matrix column is obtained by applying the sign operator and normalizing it by a factor of $\frac{1}{\sqrt{M}}$,
\[
a_j = \frac{1}{\sqrt{M}} \mathrm{sgn}(s_j).
\]
Finally, the complete sensing matrix is constructed by concatenating all generated columns,
\[
\bm{A}_{\theta} = [a_1^T, a_2^T,\dots, a_N^T]
\]
where each column is generated according to the shared rule, i.e.,
\[
a_j = \frac{1}{\sqrt{M}} \mathrm{sgn}(\mathcal{N}_\theta(z_j)).
\]
Here, the same neural network with shared parameters $\theta$ is used to generate all columns of the sensing matrix, while only the latent vector $z_j$ varies across columns. Consequently, the neural network acts as a compact matrix generator, producing a sensing matrix that exhibits a low-coherence structure.

The sensing matrix is therefore generated column-wise via a shared neural mapping, yielding a deterministic, structured construction. This approach requires only compact storage, ensures reproducibility, and provides greater mathematical control over the matrix properties. As a result, it offers significant advantages over conventional brute-force random matrix generation methods.

\subsection{Model Architecture}
\begin{figure*}[!t] 
    \centering
    \includegraphics[width= \textwidth]{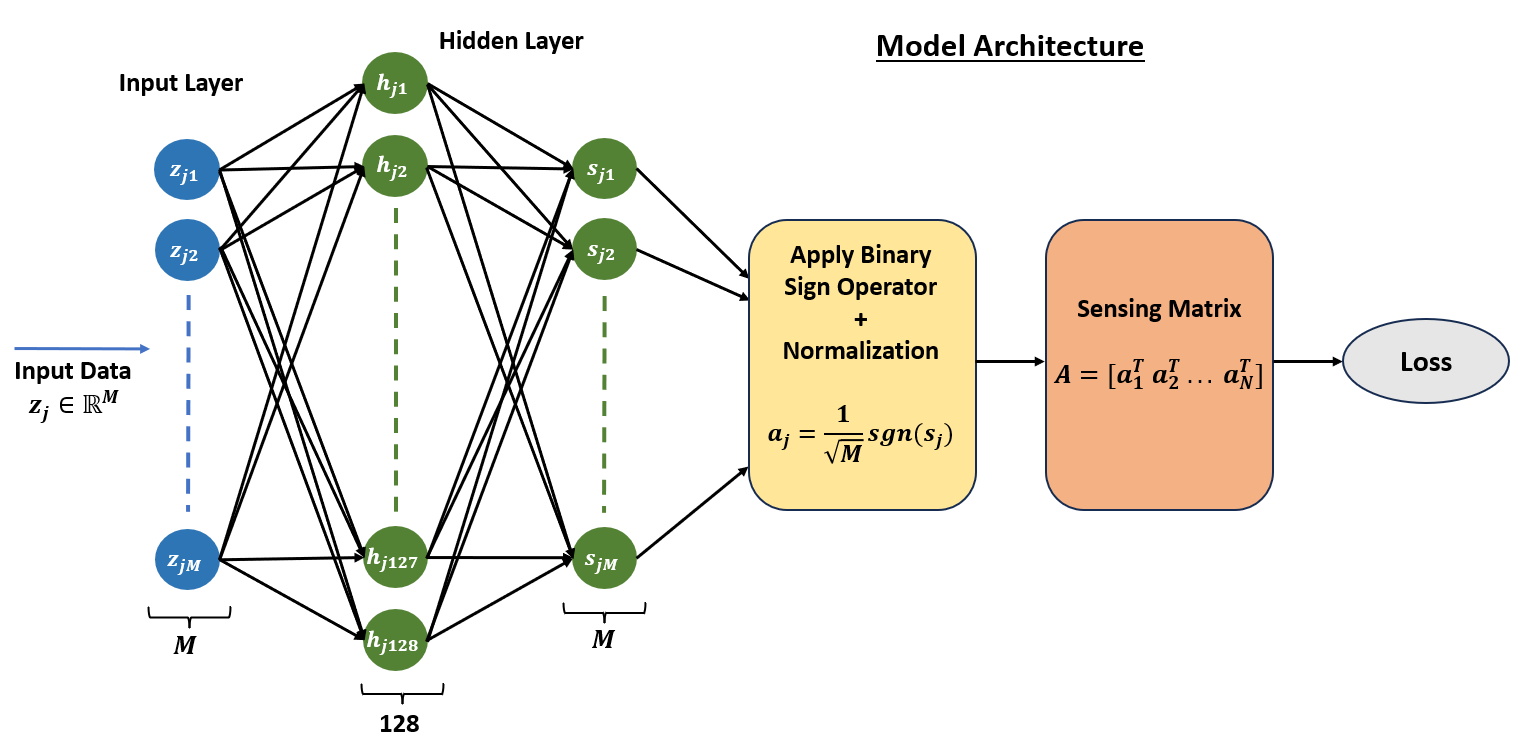}
    \caption{Architecture of the Model}
    \label{fig_6_1}
\end{figure*}

To construct the proposed property-based learned sensing matrix using a shared generation rule, a very simple fully connected feedforward neural network was employed. The architecture of the proposed model is illustrated in Figure \ref{fig_6_1}. The network consists of an input layer, two hidden layers, and an output layer.
The model takes a latent vector of dimension $M$ as the input. This input is first passed through a hidden dense layer comprising 128 neurons with the Rectified Linear Unit (ReLU) activation function. The output of this is then forwarded to another fully connected layer with 64 neurons and a linear activation function.

To enforce the binary constraint on the generated sensing matrix, the Straight-Through Estimator (STE)-based sign function is applied at the output layer. The resulting binary output is subsequently normalized by $\frac{1}{\sqrt{M}}$ to ensure stable scaling and maintain the energy normalization of the sensing matrix. Finally, the generated vectors are transposed and stacked column-wise to construct the sensing matrix of size $64 \times 256$.

\begin{figure*}[!t] 
    \centering
    \includegraphics[width= \textwidth]{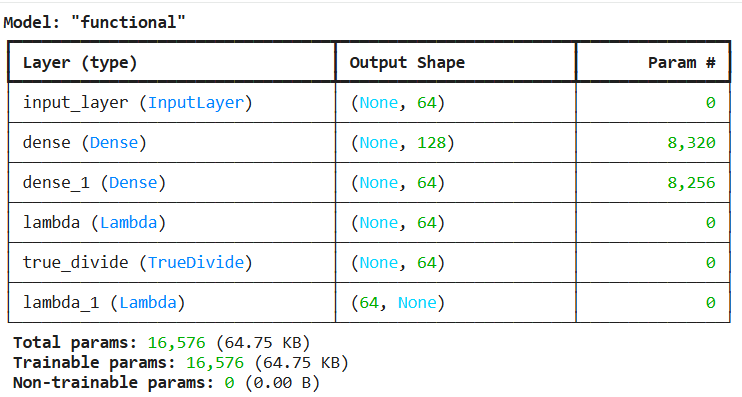}
    \caption{Summary of the Model}
    \label{fig_6_2}
\end{figure*}

The proposed model employs a simple neural network architecture rather than a complex deep network. As a result, it requires fewer trainable parameters, which reduces both computational complexity and training cost. Despite its simplicity, the model achieves competitive performance. A summary of the network architecture is presented in Figure \ref{fig_6_2}.

\subsection{Loss Function} 
In this chapter, our objective is to design a binary sensing matrix by exploiting the incoherence property rather than relying on large training datasets, thereby reducing computational complexity and training overhead. As discussed in Section \ref{sec_6_2}, the coherence among the columns of the sensing matrix $\bm{A}$ is closely related to the off-diagonal entries of its Gram matrix $\bm{G}$. Since sensing matrices with low coherence provide stronger guarantees for sparse signal recovery, minimizing coherence becomes a key design objective.

In compressive sensing, the off-diagonal elements of the Gram matrix are particularly important because they quantify the pairwise correlations among the columns of the sensing matrix. Ideally, the columns should be nearly orthogonal, resulting in a Gram matrix that approximates the identity matrix, i.e., $\bm{G} \approx \mathbb{I}$. Furthermore, the mutual coherence of $\bm{A}$ is determined by the maximum absolute off-diagonal entry of $\bm{G}$. Therefore, reducing these off-diagonal values directly improves the recovery performance of sparse signals.

However, directly minimizing the maximum off-diagonal entry is challenging because the corresponding objective function is non-differentiable and thus unsuitable for gradient-based optimization methods. To address this limitation, differentiable surrogate loss functions are commonly employed to approximate the minimization of the worst-case pairwise correlation. 

In this work, we adopt a combined loss function that simultaneously minimizes the overall coherence of the sensing matrix while penalizing the largest pairwise column correlations. This approach promotes the generation of binary sensing matrices with low mutual coherence, thereby enhancing sparse signal recovery performance. The proposed loss function is defined as
\[
\mathcal{L}(\bm{A}_{\theta})= \alpha_1 \mathcal{L}_{p-norm}(\bm{A}_{\theta})+ \alpha_2 \mathcal{L}_{LogSum} (\bm{A}_{\theta}) + \alpha_3 \mathcal{L}_{tight} (\bm{A}_{\theta})
\]
where $\alpha_1, \alpha_2 \text{ and } \alpha_3$ are weighting coefficients that control the relative contributions of the three loss components. $\mathcal{L}_{p-norm}(\bm{A}_{\theta})$, $\mathcal{L}_{LogSum} (\bm{A}_{\theta})$, and $\mathcal{L}_{tight} (\bm{A}_{\theta})$ are the $\mathcal{L}_{p}$-norm loss, LogSumExp (LSE) loss, and tight-frame loss, respectively. 

\begin{itemize}
    \item \textbf{$L_p$ norm Loss:} One of the popular choice is $L_p$ norm Loss. 
    \[
    \mathcal{L}_{p-norm} (\bm{A}_{\theta}) = \left( \sum_{i \neq j} \left|\bm{G}_{ij}\right|^p \right)^{\frac{1}{p}}
    \]
    If $p=1$, then it is equal to the sum of elements, and if $p=2$, then it becomes the Frobenius norm of $\bm{G}$. As $p \to \infty$ then the above quantity approaches to maximum value of $\bm{G}$, that is, $\mathcal{L}_{p-norm} \to \max_{i \neq j} |\bm{G}_{ij}| = \mu(\bm{A})$. 
    
    \item \textbf{LogSumExp Loss:} Instead of considering maximum absolute value of $\bm{G}$, we considered smoother approximation to the maximum. 
    \[
    \mathcal{L}_{LogSum} (\bm{A}_{\theta}) = \frac{1}{\xi} \log \left( \sum_{i \neq j} e^{\xi \bm{G}_{ij}}\right)
    \]
    As $\xi \to \infty$, this also converges to the maximum. It has a smooth gradient, which makes it numerically more stable.

    \item \textbf{Tight Frame Loss:} Since Equiangular tight frames (ETFs) can achieve the lowest possible bound of coherence, that is, the Welch bound, therefore, we consider tight frame loss. Since the tight frame satisfies $\bm{A} \bm{A}^T = \frac{N}{M} \mathbb{I}$, therefore we cosider the loss function 
    \[
    \mathcal{L}_{tight} (\bm{A}_{\theta}) = \|\bm{A} \bm{A}^T - \frac{N}{M} \mathbb{I}\|_F^2
    \]
    where $\|.\|_F$ denotes the Frobenius norm.
\end{itemize}

By appropriately selecting the weighting coefficients, the optimization process can effectively balance reducing the average column correlation with suppressing the maximum pairwise correlation. Consequently, the proposed loss function directly targets the coherence properties of the sensing matrix, guiding the learning process toward constructing binary sensing matrices with improved incoherence and enhanced sparse recovery capability.

\begin{algorithm}[h!]
\caption{: Property-driven learned binary sensing matrix construction}\label{algo_6_1}
\begin{algorithmic}[0]
    \State \textbf{Input:} $N$ latent vector $z_j \in \mathbb{R}^M$ sampled from some distribution, number of epochs$(E)$, learning rate$(l)$, weighting coefficients of the loss function $\alpha_1, \alpha_2, \text{ and }, \alpha_3$, loss function parameters $p$ and $\xi$.
    \State \textbf{Output:} Sensing matrix $\bm{A}\in \mathbb{R}^{M \times N}$.
\end{algorithmic}

\begin{algorithmic}[0]
        \State \vspace{1px}
        \State 1. \textbf{Initialization:} Randomly initialize weights and biases of the network.
        \State \vspace{1px}
        \State 2. \textbf{Forward Propagation:} 
        \State \hspace{1em} a). Feed the latent vectors to the input layer with $M$ neurons.
        \State \hspace{1em} b). Calculate the value of hidden layer neurons, i.e.,
        \[H=A_1(ZW_1 + b_1)\]
        \[S=A_2(HW_2 + b_2)\]
        where $A_1$ and $A_2$ are the activation functions of the hidden layers.
        \State \hspace{1em} c). \textbf{Binary Constraint and Scale: } Apply binary constraint and scale  i.e.,
        \[A_{in} = \frac{1}{\sqrt{M}}\mathrm{sgn}(S)\]
        \State \hspace{1em} d). \textbf{Sensing Matrix Construction: } The sensing matrix $\bm{A} = A_{in}^T$.
        \State \vspace{1px}
        \State 3. \textbf{Compute Loss: } Calculate the loss by
        \[
        \mathcal{L}(\bm{A})= \alpha_1 \mathcal{L}_{p-norm}(\bm{A})+ \alpha_2 \mathcal{L}_{LogSum} (\bm{A}) + \alpha_3 \mathcal{L}_{tight} (\bm{A})
        \]
        where $\mathcal{L}_{p-norm}(\bm{A})$, $\mathcal{L}_{LogSum}(\bm{A})$ and $\mathcal{L}_{tight}(\bm{A})$ are calculated as describe in section \ref{sec_6_5}.
        \State \vspace{1px}
        \State 4. \textbf{Back Propagation: } Compute gradient and update the network parameters using Adam optimizer. 
        \State \vspace{1px}
        \State To learn the optimal parameters, repeat steps 2 to 4, $E$ number of times.
    \State \vspace{1px}
    \State \textbf{Return} $\bm{A}$
\end{algorithmic}
\end{algorithm}

\section{Simulation Result of Proposed Model} \label{sec_6_6}
This section presents simulation results for the proposed learning-based method that learns a binary sensing matrix satisfying the incoherence property. The most important thing is to train our model; we don't need a large amount of data. We feed our model a latent vector sample from a Gaussian distribution, and by learning from the incoherence property and following the common generation rule, it generates incoherent columns of the sensing matrix.

\subsection{Training Data} \label{sec_6_6_1}
To train the model, we don't use a large amount of data. In fact, our model learns from the classical CS incoherence property rather than from data. Most learning-based methods use data for training; as a result, they learn data statistics, and their performance is mostly data-specific, leading to limited generalization. 

In our model, instead of optimizing each entry of $\bm{A}$, we optimize it column-wise. To generate the columns of the sensing matrix, we feed $N$ latent vectors of dimension $M$ sampled from a Gaussian distribution. And our model maps these latent vectors to the columns of the sensing matrix $\bm{A}$.

\subsection{Hyperparameter Tuning} \label{sec_6_6_2}
To determine the optimal hyperparameters, extensive experiments were conducted by varying values of hyperparameters, including learning rate, the number of neurons in the hidden layer, the number of training epochs, and the weighting coefficients of the proposed loss function, namely $\alpha_1, \alpha_2, \text{ and } \alpha_3$. A grid search strategy was used to systematically explore different parameter combinations. During the search process, both the total loss $\mathcal{L}(\cdot)$ and the maximum $\mu(\cdot)$ mutual coherence were monitored, since the primary objective of the proposed approach is to minimize the maximum mutual coherence of the resulting sensing matrix.

Although several hyperparameter combinations yielded comparable total loss values, the final configuration was selected for its ability to simultaneously minimize both the total loss and the maximum mutual coherence. In particular, a hidden layer consisting of $128$ neurons was adopted. While increasing the number of neurons to 256 resulted in a marginal reduction in total loss, it did not yield a noticeable improvement in the maximum mutual coherence. Therefore, the simpler architecture with 128 neurons was preferred due to its lower computational complexity.

The learning rate was set to 0.0001, and the weighting coefficients for the loss function were set to $\alpha_1 = 3, \alpha_2 = 1, \text{ and } \alpha_3 = 0.5$. Furthermore, the parameters $p = 8$ and $\xi = 30$ were selected because larger values cause $\mathcal{L}_{p-norm}(\cdot)$ and $\mathcal{L}_{LogSum}(\cdot)$ to more closely approximate the maximum mutual coherence. Using this hyperparameter configuration, the proposed model was trained for 1000 epochs.

\subsection{Complexity of Model}
To construct the random sensing matrix of size $M \times N$, the construction and storage complexity will be $O(MN)$, which may not seem much, but the structure of the random matrix makes it complex for further calculation, as the sensing matrix is the one that is used for signal acquisition as well as for reconstruction. These processes involve extensive calculations and matrix multiplication, making the system computationally intensive. 

Most existing learning-based matrices rely on very complex networks and large amounts of training data. Because of this, not only the construction complexity but also the storage complexity increases, as they depend on the network's parameters, which are generally very large.

However, in our case, we used a very small fully connected neural network, and its storage complexity is almost equal to that of the random sensing matrix. After training the model, we just need to store the network parameters to construct a new sensing matrix, which provides much better incoherence guarantees than the random matrix. The matrix's binary nature makes it much easier to handle and perform further calculations.

\subsection{Model Performance and Comparison Result} \label{sec_6_6_3}
To design a binary sensing matrix with low mutual coherence, we proposed a neural network framework using the coherence-based loss function described in Section \ref{sec_6_5}. For the selected hyperparameters described in section \ref{sec_6_6_2}, the model is trained for 1000 epochs across different sensing matrix dimensions $(M, N)$. The convergence behavior of the proposed approach is illustrated through several coherence-related performance measures.

\begin{figure*}[!t]
\centering
\subfloat[$M=64,N=128$]{\includegraphics[width=0.32\textwidth]{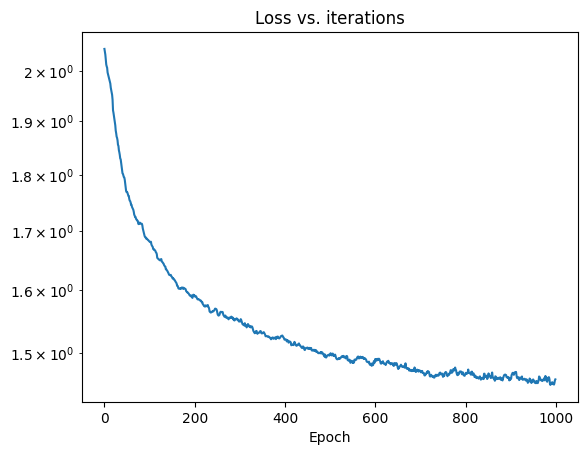}}
\label{fig_6_3_a}
\hfill
\subfloat[$M=64,N=256$]{\includegraphics[width=0.32\textwidth]{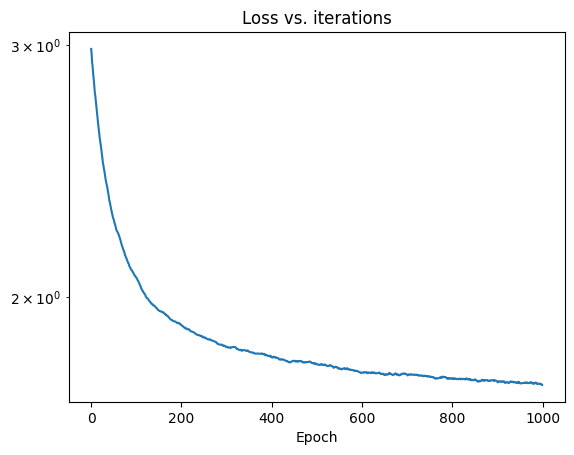}}
\label{fig_6_3_b}
\hfill
\subfloat[$M=128,N=256$]{\includegraphics[width=0.32\textwidth]{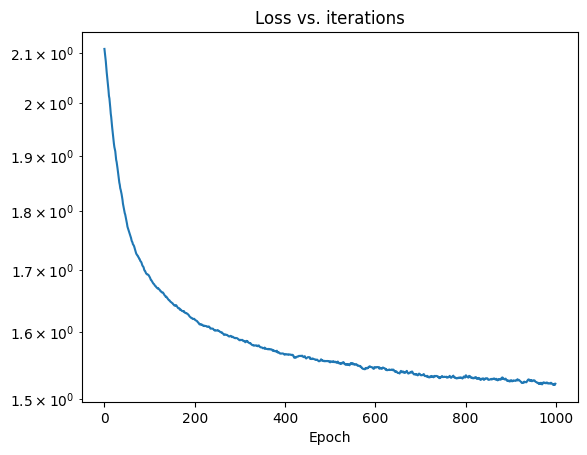}}
\label{fig_6_3_c}
\caption{Loss plot versus iteration for different sensing matrix dimensions $(M,N)$}
\label{fig_6_3}
\end{figure*}

Figure \ref{fig_6_3} presents the convergence of the loss function over the varying iterations. A consistent decrease in loss value is observed, demonstrating the effectiveness of the optimization process and indicating that the learned sensing matrix progressively approaches the desired low-coherence structure. 

\begin{figure*}[!t]
\centering
\subfloat[$M=64,N=128$]{\includegraphics[width=0.32\textwidth]{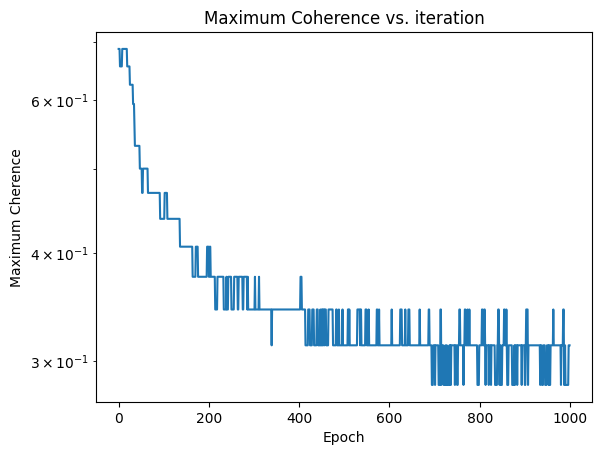}}
\label{fig_6_4_a}
\hfill
\subfloat[$M=64,N=256$]{\includegraphics[width=0.32\textwidth]{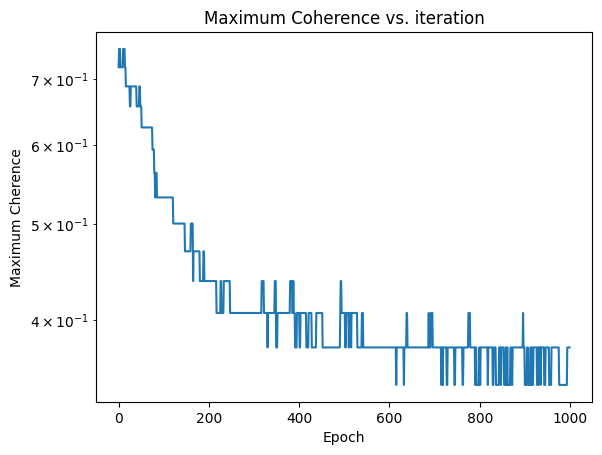}}
\label{fig_6_4_b}
\hfill
\subfloat[$M=128,N=256$]{\includegraphics[width=0.32\textwidth]{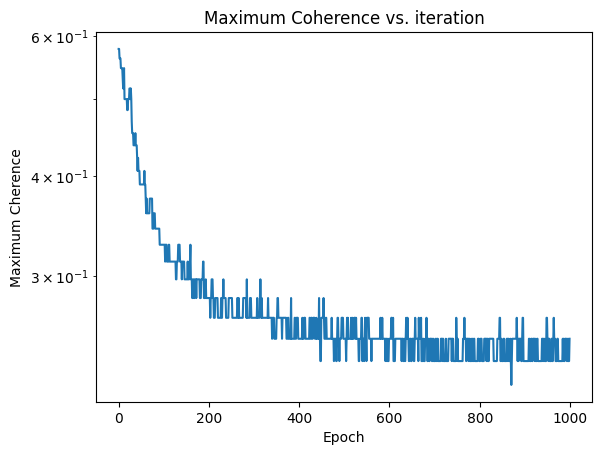}}
\label{fig_6_4_c}
\caption{Maximum mutual coherence versus iteration for different sensing matrix dimensions $(M,N)$}
\label{fig_6_4}
\end{figure*}

Figure \ref{fig_6_4} shows the variation in the maximum mutual coherence over iterations. However, the maximum coherence does not decrease monotonically with iterations and exhibits local fluctuations. This reflects the switching of the worst-case coherence pair across iterations. Despite these fluctuations, the envelope of the curve decreases progressively, indicating that the overall coherence structure of the proposed sensing matrix is improving.

\begin{figure*}[!t]
\centering
\subfloat[$M=64,N=128$]{\includegraphics[width=0.32\textwidth]{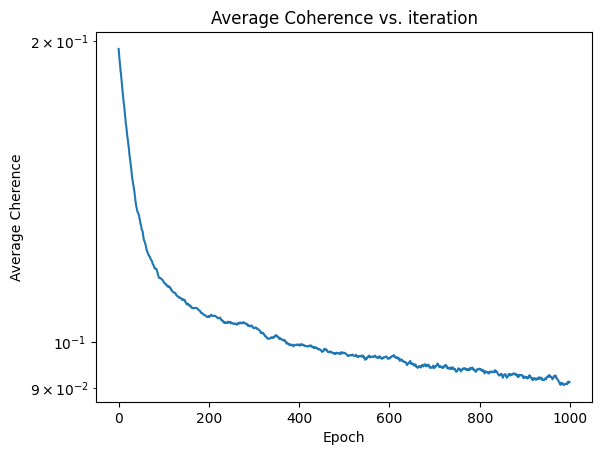}}
\label{fig_6_5_a}
\hfill
\subfloat[$M=64,N=256$]{\includegraphics[width=0.32\textwidth]{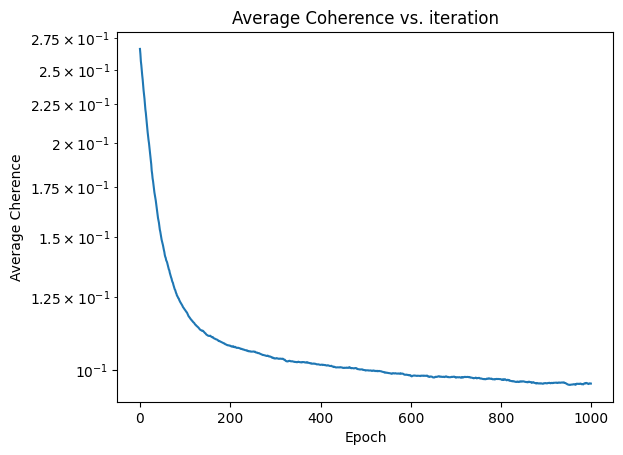}}
\label{fig_6_5_b}
\hfill
\subfloat[$M=128,N=256$]{\includegraphics[width=0.32\textwidth]{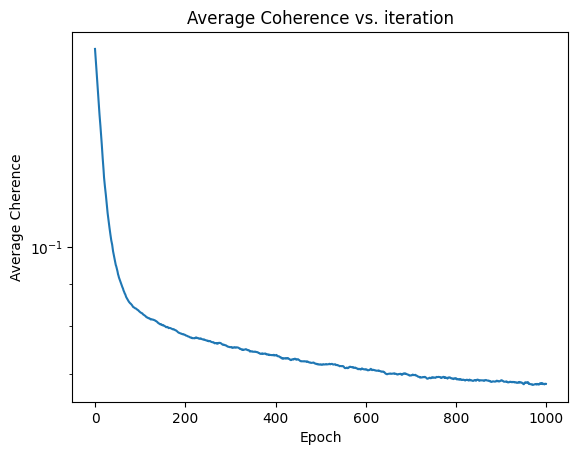}}
\label{fig_6_5_c}
\caption{Average mutual coherence versus iteration for different sensing matrix dimensions $(M,N)$}
\label{fig_6_5}
\end{figure*}

\begin{figure*}[!t]
\centering
\subfloat[$M=64,N=128$]{\includegraphics[width=0.32\textwidth]{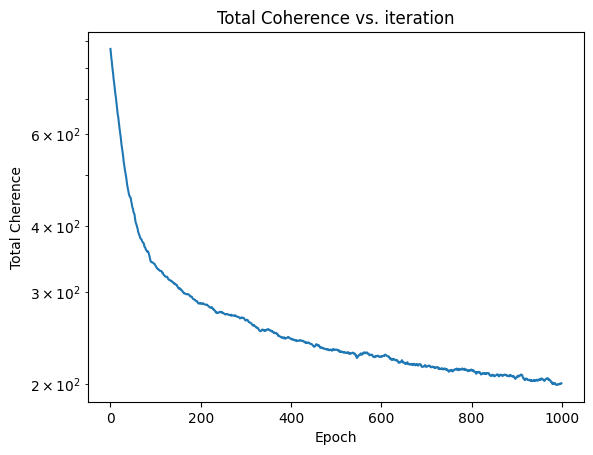}}
\label{fig_6_6_a}
\hfill
\subfloat[$M=64,N=256$]{\includegraphics[width=0.32\textwidth]{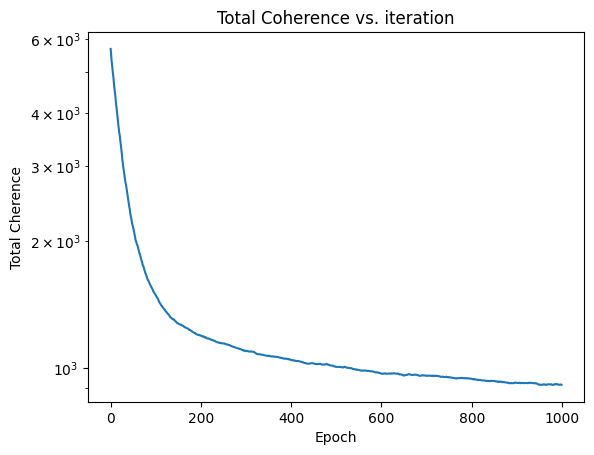}}
\label{fig_6_6_b}
\hfill
\subfloat[$M=128,N=256$]{\includegraphics[width=0.32\textwidth]{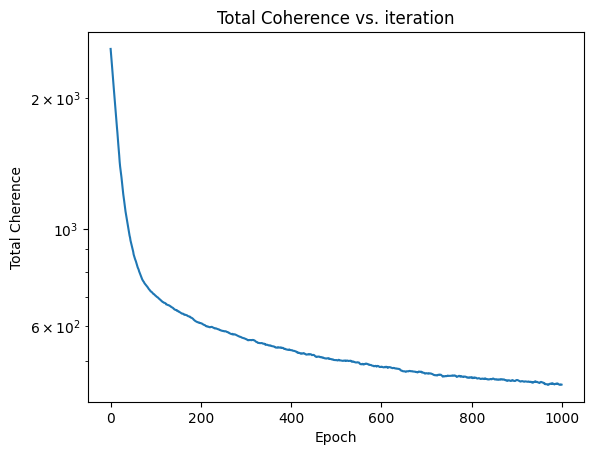}}
\label{fig_6_6_c}
\caption{Total mutual coherence versus iteration for different sensing matrix dimensions $(M,N)$}
\label{fig_6_6}
\end{figure*}

In contrast, average and total mutual coherence are less sensitive and exhibit smoother, strictly decreasing trends as shown in Figures \ref{fig_6_5} and \ref{fig_6_6}, respectively. This further validates the capability of the proposed learning strategy to reduce overall column correlations.

\begin{figure*}[!t]
\centering
\subfloat[$M=64,N=128$]{\includegraphics[width=0.32\textwidth]{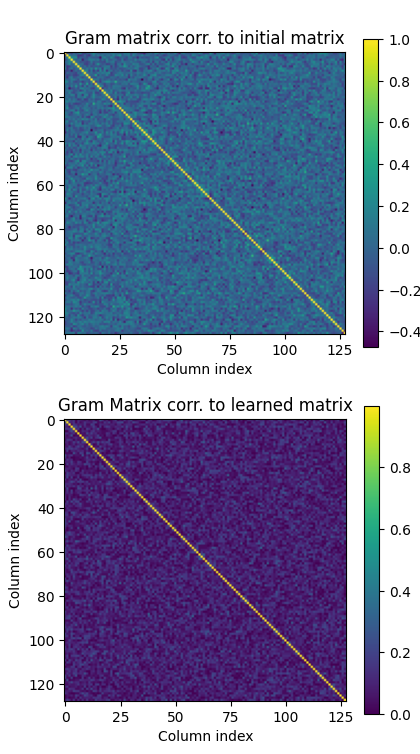}}
\label{fig_6_7_a}
\hfill
\subfloat[$M=64,N=256$]{\includegraphics[width=0.32\textwidth]{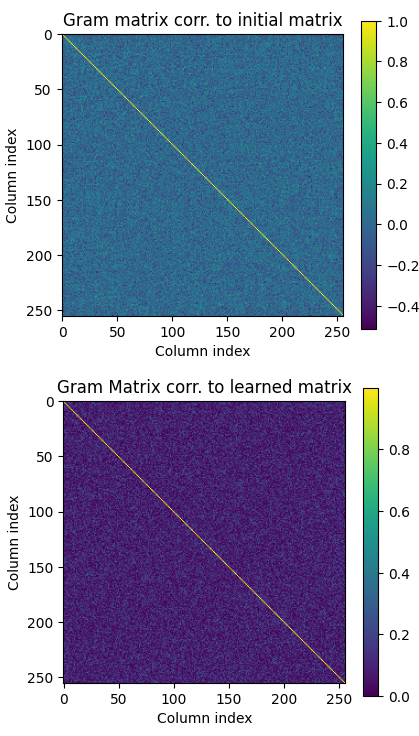}}
\label{fig_6_7_b}
\hfill
\subfloat[$M=128,N=256$]{\includegraphics[width=0.32\textwidth]{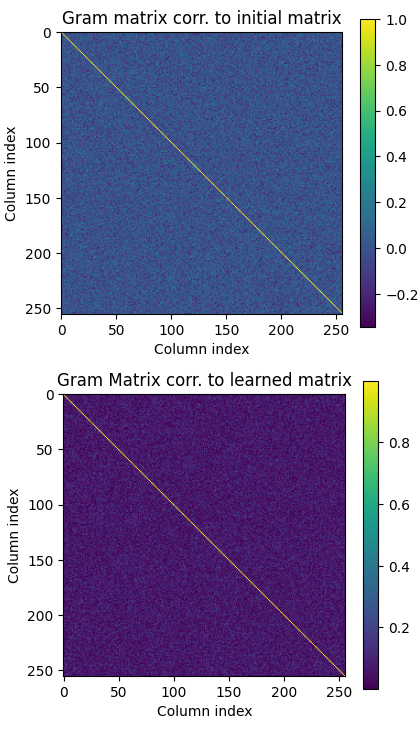}}
\label{fig_6_7_c}
\caption{Comparison of the heatmap of the Gram matrix entries corresponding to the initial sensing matrix and the proposed learned sensing matrix for different sensing matrix dimensions $(M, N)$.}
\label{fig_6_7}
\end{figure*}

To provide a visual interpretation of the learned matrix structure, Figure \ref{fig_6_7} compares the heatmaps of the Gram matrices corresponding to the initial sensing matrix and the learned sensing matrix for different values of $(M, N)$. The first row shows the initial Gram matrices, while the second row presents the Gram matrices obtained after optimization. The learned matrices exhibit significantly weaker off-diagonal entries, indicating reduced inter-column correlations and improved incoherence properties.

Furthermore, we compare the proposed method with two commonly used random sensing matrices: (i) a Gaussian random matrix whose entries are independently drawn from a normal distribution with mean zero and variance one, and (ii) a Bernoulli random matrix whose entries take values $\pm \frac{1}{\sqrt{M}}$ with equal probability. The coherence characteristics of these matrices are compared with those of the proposed learned sensing matrix.

Table \ref{tab_6_1} presents a comparison of the maximum, average, and total mutual coherence values achieved by the proposed learned sensing matrix and randomly generated sensing matrices for different matrix dimensions $(M, N)$. For the proposed approach, the reported values correspond to the averages obtained from 20 independent runs of the learning algorithm. For the random matrices, 20 were generated for each M and N case, and the maximum, average, and total coherence were computed for each matrix; the values reported in the table represent the corresponding mean value. As shown, the proposed learned sensing matrix consistently yields lower coherence measures across all considered matrix dimensions. These results demonstrate the effectiveness of the proposed learning framework in constructing highly incoherent sensing matrices, which are desirable for improved compressed sensing performance.
\begin{table}[htbp]
\centering
\caption{Comparison of mutual coherence and related performance measures of the proposed learned sensing matrix and random sensing matrices (Gaussian and Bernoulli matrices) for different values of $M$ and $N$.\label{tab_6_1}}
\centering
\begin{tabularx}{\textwidth}
{|c|>{\centering\arraybackslash}X|
 >{\centering\arraybackslash}X|
 >{\centering\arraybackslash}X|}
\hline
\multicolumn{4}{| c |}{\textbf{$M=64$ and $N = 128$}}\\
\hline
\textbf{Matrix} & \textbf{Maximum Mutual Coherence ($\mu(\bm{A})$)} & \textbf{Average Mutual Coherence ($\mu_{avg}(\bm{A})$)} & \textbf{Total Mutual Coherence ($\mu_{total}(\bm{A})$)}\\
\hline
Random Gaussian & $\approx 0.481$ & $\approx 0.099$ & $\approx 256.23$ \\

Random Bernoulli & $\approx 0.496$ & $\approx 0.098$ & $\approx 253.34$ \\

Proposed & $\approx \textbf{0.281}$ & $\approx \textbf{0.091}$ & $\approx \textbf{200.58}$ \\
\hline
\multicolumn{4}{| c |}{\textbf{$M=64$ and $N = 256$}}\\
\hline
Random Gaussian & $\approx 0.501$ & $\approx 0.099$ & $\approx 1022.83$ \\

Random Bernoulli & $\approx 0.528$ & $\approx 0.098$ & $\approx 1019.58$ \\

Proposed & $\approx \textbf{0.343}$ & $\approx \textbf{0.092}$ & $\approx \textbf{854.37}$ \\
\hline
\multicolumn{4}{| c |}{\textbf{$M=128$ and $N = 256$}}\\
\hline
Random Gaussian & $\approx 0.368$ & $\approx 0.070$ & $\approx 510.52$ \\

Random Bernoulli & $\approx 0.368$ & $\approx 0.070$ & $\approx 509.27$ \\

Proposed & $\approx \textbf{0.250}$ & $\approx \textbf{0.068}$ & $\approx \textbf{441.9 7}$ \\
\hline
\end{tabularx}
\end{table}

\begin{figure*}[!t]
\centering
\subfloat[$M=64,N=128$]{\includegraphics[width=0.32\textwidth]{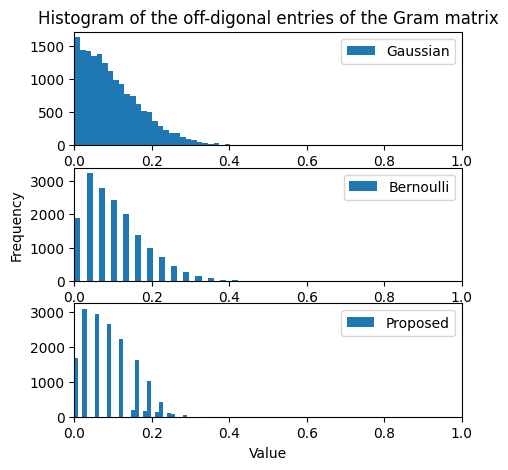}}
\label{fig_6_8_a}
\hfill
\subfloat[$M=64,N=256$]{\includegraphics[width=0.32\textwidth]{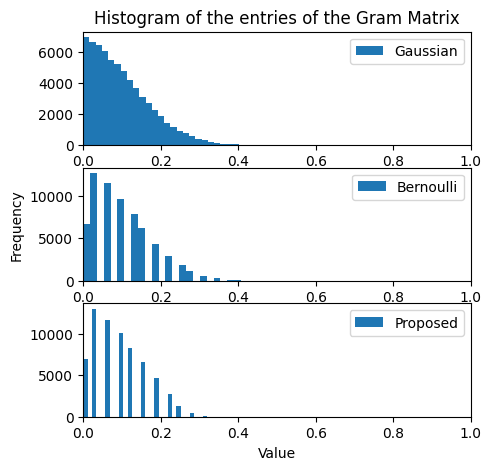}}
\label{fig_6_8_b}
\hfill
\subfloat[$M=128,N=256$]{\includegraphics[width=0.32\textwidth]{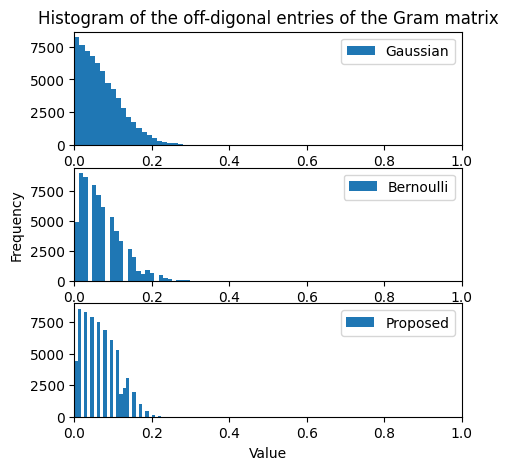}}
\label{fig_6_8_c}
\caption{Comparison of the coefficient distributions of the off-diagonal Gram entries of the proposed learned sensing matrix and existing random sensing matrices (Gaussian and Bernoulli) through histogram analysis for different sensing matrix dimensions $(M, N)$.}
\label{fig_6_8}
\end{figure*}

Figure \ref{fig_6_8} presents the histograms of the off-diagonal entries of the corresponding Gram matrices for different values of $(M,N)$. The histogram associated with the proposed learned sensing matrix is more concentrated around zero and exhibits considerably shorter tails than those of the random matrices. This behavior indicates that the correlations between distinct columns are substantially reduced. In particular, the largest-magnitude off-diagonal entries are significantly smaller for the proposed method, which directly translates into lower maximum mutual coherence.

Overall, the experimental results, including the convergence analysis, Gram matrix visualization, histogram comparisons, and quantitative evaluations reported in Table \ref{tab_6_1}, demonstrate that the proposed coherence-driven learning framework effectively constructs sensing matrices with substantially reduced maximum, average, and total mutual coherence. These findings confirm the superiority of the proposed method over conventional random-sensing-matrix constructions across a wide range of matrix dimensions.

\section{Conclusion}
In compressive sensing, the sensing matrix plays a crucial role in acquiring and accurately reconstructing signals from a limited number of measurements. Over the years, several methods for constructing sensing matrices have been proposed in the literature. However, many of these methods suffer from limitations such as high computational complexity, complicated construction procedures, or failure to satisfy the essential properties required for effective compressive sensing.

To address these challenges, this chapter proposed a learning-based approach for constructing a binary sensing matrix using an artificial neural network (ANN). Unlike conventional data-driven methods, the proposed network learns directly from the mutual incoherence property, a key characteristic of an effective sensing matrix.

A simple neural network with only a few layers was designed to optimize the sensing matrix. The objective was to minimize the matrix's mutual coherence. However, since the maximum mutual coherence is not differentiable, it cannot be used directly as a loss function. Therefore, differentiable surrogate functions were employed, and a combined loss function was designed to reduce the overall correlation among the matrix columns while encouraging low maximum coherence.

The effectiveness of the proposed method was evaluated by comparing its coherence measures with those of randomly generated sensing matrices. In addition, the corresponding Gram matrices were analyzed to assess the correlation structure. The experimental results and visual comparisons demonstrate that the proposed approach produces sensing matrices with favorable coherence properties and achieves competitive performance.

\clearpage
\bibliographystyle{plain}
\bibliography{reference}
\end{document}